\pdfoutput=1

\documentclass[11pt]{article}

\usepackage{acl}

\usepackage{times}
\usepackage{latexsym}
\usepackage[T1]{fontenc}
\usepackage[utf8]{inputenc}
\usepackage{microtype}
\usepackage{inconsolata}

\usepackage{amsmath,amssymb}
\usepackage{graphicx}
\usepackage{subcaption}
\usepackage{booktabs}
\usepackage{paralist}
\usepackage{accents}
\usepackage{float}
\usepackage[nameinlink,capitalise]{cleveref}
\usepackage{caption}
\usepackage{xcolor}

\usepackage{algorithm}
\usepackage[noend]{algorithmic}

\usepackage[T1]{fontenc}

\usepackage[utf8]{inputenc}

\usepackage{microtype}

\usepackage{inconsolata}

\usepackage{tikz}
\usepackage{tikz-qtree}
\usepackage{forest}
\usetikzlibrary{calc,backgrounds,bayesnet,positioning,fit,patterns,matrix}
\tikzstyle{every picture}+=[remember picture]
\tikzstyle{-|}=[to path={-| (\tikztotarget)}]
\tikzstyle{|-}=[to path={|- (\tikztotarget)}]

\crefname{section}{\S}{\S\S}
\Crefname{section}{\S}{\S\S}
\crefformat{equation}{(#2#1#3)}
\DeclareMathOperator{\R}{{}\mathbb{R}}   
\newcommand{\upe}{\bar{e}}
\newcommand{\downe}{\underaccent{\bar}{e}}
\newcommand{\pemb}{\Psi}
\newcommand{\pcomp}{\Phi}
\newcommand{\pdecomp}{\Theta}

\title{Self-StrAE at SemEval-2024 Task 1: Making Self-Structuring AutoEncoders  Learn More With Less }

\author{%
  Mattia Opper\textsuperscript{\,a}
  \and N. Siddharth\textsuperscript{\,a,b} \\[0.5ex]
  \textsuperscript{a} University of Edinburgh;\;
  \textsuperscript{b} The Alan Turing Institute \\[0.5ex]
  \texttt{\{m.opper,n.siddharth\}@ed.ac.uk}
}

\begin{document}
\maketitle
\begin{abstract}
This paper presents two simple improvements to the Self-Structuring AutoEncoder (Self-StrAE). Firstly, we show that including reconstruction to the vocabulary as an auxiliary objective improves representation quality. Secondly, we demonstrate that increasing the number of independent channels leads to significant improvements in embedding quality, while simultaneously reducing the number of parameters. Surprisingly, we demonstrate that this trend can be followed to the extreme, even to point of reducing the total number of non-embedding parameters to seven. Our system can be pre-trained from scratch with as little as 10M tokens of input data, and proves effective across English, Spanish and Afrikaans. 
\end{abstract}

\section{Introduction}

Natural language is generally understood to be compositional. To understand a sentence, all you need to know are the meanings of the words and how they fit together. The mode of combination is generally conceived as an explicitly structured hierarchical process which can be described through, for example, a parse tree. Recent work by \citet{strae} presents the Self-StrAE (Self-Structuring AutoEncoder), a model which learns embeddings such that they define their own hierarchical structure and extend to multiple levels (i.e. from the subword to the sentence level and beyond). The strengths of this model lie in its parameter and data efficiency achieved through the inductive bias towards hierarchy. 

Learning embeddings such that they meaningfully represent semantics is crucial for many modern NLP applications. For example, retrieval augmented generation \cite{RAG} is predicated on the fact that the correct contexts for a given query can be determined. The semantic relation between a query and a context is encompassed by the notion of semantic relatedness. They are not equivalent to one another (i.e. paraphrases), but are close in meaning in a broader, more contextual sense. The focus of task one of this year's SemEval \cite{semrel, semreltask} is capturing this notion of semantic relatedness, with a particular focus on African and Asian languages generally characterised by a lack of NLP resources. 

In this work, we investigate whether Self-StrAE can learn embeddings which capture semantic relatedness, when trained from scratch on moderately sized pre-training corpora. We turn to the competition in order to examine whether the model can even compare with dedicated STR systems. In order to determine whether Self-StrAE can provide an alternative approach in low resource settings where systems that rely on large pre-trained transformers \cite{citationsisallyouneed} may not have sufficient scale to prove effective.  We show that with two simple changes, Self-StrAE's performance can be substantially improved. Moreover, we demonstrate that the the resulting system is not limited to English, but can work equally well (if not better) for both Spanish and Afrikaans \footnote{Code available at: \url{https://github.com/mopper97/Self-StrAE}}. 

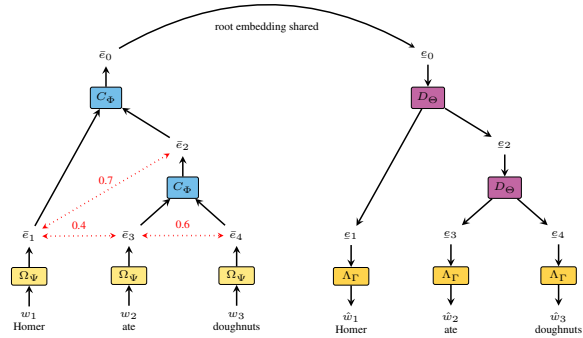
\begin{figure}[t]
  \centering
  \resizebox{\columnwidth}{!}{%
  \begin{tikzpicture}[
    thick,>=stealth,font=\tiny,
    every tree node/.style={align=center,anchor=north,inner sep=3pt},
    edge from parent/.append style={<-,thick},
    edge from parent path={(\tikzparentnode) -- (\tikzchildnode);},
    level 1+/.style={level distance=24pt,sibling distance=32pt},
    fnb/.style={draw,rounded corners=1pt,yshift=2pt},
    comp/.style={fnb,fill=cyan!80!blue!50},
    decomp/.style={fnb,fill=purple!80!blue!60},
    emb/.style={fnb,inner sep=2pt,text width=3ex,fill=yellow!80!orange!60},
    demb/.style={fnb,inner sep=2pt,text width=3ex,fill=yellow!60!orange!80},
    lf/.style={yshift=5pt},
    ]
    \Tree [.\node(ur){\(\upe_0\)};
            [.\node[comp]{\(C_{\pcomp}\)};
              [.\node[yshift=-1.7cm](e1){\(\upe_1\)};
                [.\node[emb,yshift=-1.7cm]{\(\Omega_{\pemb}\)};
                  [.\node[lf,yshift=-1.7cm]{\(w_1\)\\Homer};]]]
              [.\node(e2){\(\upe_2\)};
                [.\node[comp]{\(C_{\pcomp}\)};
                  [.\node(e3){\(\upe_3\)};
                    [.\node[emb]{\(\Omega_{\pemb}\)};
                      [.\node[lf]{\(w_2\)\\ate};]]]
                  [.\node(e4){\(\upe_4\)};
                    [.\node[emb]{\(\Omega_{\pemb}\)};
                      [.\node[lf]{\(w_3\)\\doughnuts};]]]]]]]
  \draw[<->,red,dotted] (e1) -- (e2) node[midway, above] {0.7};
  \draw[<->,red,dotted] (e1) -- (e3) node[midway, above] {0.4};
  \draw[<->,red,dotted] (e3) -- (e4) node[midway, above] {0.6};
  \begin{scope}[shift={(6cm,0)}]
  \tikzset{edge from parent/.append style={->,thick}}
  \Tree [.\node(dr){\(\downe_0\)};
        [.\node[decomp]{\(D_{\pdecomp}\)};
          [.\node[yshift=-1.7cm](de1){\(\downe_1\)};
            [.\node[demb,yshift=-1.7cm]{\(\Lambda_{\Gamma}\)};
              [.\node[lf,yshift=-1.7cm]{\(\hat{w}_1\)\\Homer};]]]
          [.\node(de2){\(\downe_2\)};
            [.\node[decomp]{\(D_{\pdecomp}\)};
              [.\node(de3){\(\downe_3\)};
                [.\node[demb]{\(\Lambda_{\Gamma}\)};
                  [.\node[lf]{\(\hat{w}_2\)\\ate};]]]
              [.\node(de4){\(\downe_4\)};
                [.\node[demb]{\(\Lambda_{\Gamma}\)};
                  [.\node[lf]{\(\hat{w}_3\)\\doughnuts};]]]]]]]
  \end{scope}
  \draw[->] (ur) to[bend left=30] node[midway,below,yshift=-2mm] {root embedding shared}    (dr);
  \end{tikzpicture}
  }
  \caption{Self-StrAE forward pass. Red lines indicate cosine similarity between adjacent nodes. Shared colours indicate shared parameters.}
  \label{fig: selfstrae}
\end{figure}

\newpage

\section{Model and Objectives}

\subsection{Model}

The core architecture at the heart of this paper is the Self-StrAE. A model that processes a given sentence to generate both multi-level embeddings and a structure over the input. The forward pass begins by first \textit{embedding} tokens to form an initial frontier, using the embedding matrix \(\Omega_{\pemb}\). This is followed by iterative application of the following update rule:

\begin{enumerate}
    \item Take the cosine similarity between adjacent embeddings in the frontier. 
    \item Pop the most similar pair. 
    \item Merge the pair into a single parent representation, and insert into the frontier. 
    \item If len(frontier) = 1, stop
\end{enumerate}

Merge is handled by the recursively applied \textit{composition function} \(C_{\pcomp}\), which takes the embeddings of two children and produces that of the parent. The process is illustrated in \ref{fig: selfstrae}. In the figure, the highest cosine similarity is between the embeddings of 'ate' and 'doughnuts', so these two embeddings are merged first. At the next step, 'Homer' and 'ate doughnuts' are merged as they have the highest similarity of the remaining embeddings. At this point the frontier has shrunk to a single embedding and the root has been reached.

If we consider the merge history at the root, we can see that it has come to define a tree structure over the input. This structure is passed to the decoder, which then generates a second set of embeddings, starting from the root and proceeding to the leaves. The decoder achieves this through recursive application of the \textit{decomposition function} \(D_{\pdecomp}\), which takes the embedding of a parent and produces the embeddings of the two children. Once the decoder reaches the leaves, it can optionally output discrete tokens through use of a \textit{dembedding function} \(\Lambda_{\Gamma}\). 

We denote embeddings produced during composition as $\upe$ and produced during decomposition as $\downe$. For a vocabulary of size \textit{V}, each embedding $e \in \mathbb{R}^{E}$ consists of \textit{k} independent channels of size \textit{u}. With this notation established, we can now define the four core components of a Self-StrAE. \\
\newline 
\newline \newline
\noindent\textbf{Embedding:}

$\Omega_{\pemb} (w_{i}) = w_{i}\pemb$, where $\pemb \in \mathbb{R}^{V \times E}$ \\

\noindent\textbf{Composition:}

$C_{\pcomp} (\upe_{c1}, \upe_{c2}) = hcat(\upe_{c1}, \upe_{c2})\pcomp + \phi$ 

where $\pcomp \in \mathbb{R}^{2u \times u}$ and $\phi \in \mathbb{R}^{u}$\\

\noindent\textbf{Decomposition:}

$D_{\pdecomp}(\downe_{p}) = hsplit (\downe_{p}\pdecomp + \theta)$ 

where $\pdecomp \in \mathbb{R}^{u \times 2u}$ and $\theta \in \mathbb{R}^{2u}$\\

\noindent\textbf{Dembedding:}

$\Lambda_{\Gamma}(\downe_{i}) = \downe_{i}\Gamma$ where $\Gamma \in \mathbb{R}^{E \times V}$ \\

Note that in the above the dembedding layer is treated as a separate parameter matrix to the embedding layer, however, it can just as easily be weight tied to increase efficiency.

\subsection{Objectives}
There are a few options for pre-training Self-StrAE. The simplest solution is to have the model reconstruct the leaf tokens, which can be achieved by simply employing cross entropy over the output of the dembedding layer. For a given sentence~\(s_j = \langle w_i \rangle_{i=1}^{T_j}\), this objective is formulated as:
\begin{align}
  \label{eq:ce-loss}
  \mathcal{L}_\text{CE} = - \frac{1}{T_j} \sum_{i=1}^{T_j} w_{i} \cdot \log \hat{w}_{i}.
\end{align}






An alternative approach adopted by \citet{strae} is to use contrastive loss as the reconstruction objective. For a given batch of sentences~\(s_j\), the total number of nodes (internal + leaves) in the associated structure is denoted as~\(M\).
This allows for the construction of a pairwise similarity matrix~\(A \in \R^{M \times M}\) between normalised upward embeddings~\(\langle \upe_i \rangle_{i=1}^M\) and normalised downward embeddings~\(\langle \downe_i \rangle_{i=1}^M\), using the cosine similarity metric (where embeddings are flattened to be of shape \textit{E}).
Denoting \(A_{i\bullet}, A_{\bullet{}j}, A_{ij}\) the \(i\textsuperscript{th}\) row, \(j\textsuperscript{th}\) column, and \((i,j)\textsuperscript{th}\) entry of a matrix respectively, the objective is defined as:
\begin{align}
  \label{eq:cont-obj}
  \mathcal{L}_{\text{cont}}
  = \frac{-1}{2M}\!\!\left[\!
  \sum_{i=1}^M \log \sigma_{\!\tau}(A_{i\bullet}) \!+\! \sum_{j=1}^M \log \sigma_{\!\tau}(A_{\bullet{}j})
  \!\right]\!
\end{align}
where~\(\sigma_{\!\tau}(\cdot)\) is the tempered \texttt{softmax} (temperature~\(\tau\)), normalising over the unspecified (\({}_\bullet\)) dimension.

A final option is to combine these two objectives, applying the cross entropy reconstruction over leaves and the contrastive objective over all other nodes, where constructing a vocabulary is intractable due to the number of possible combinations. The contrastive objective remains identical except that \textit{A} is now defined as pairwise similarity matrix~\(A \in \R^{I \times I}\), where \textit{I} is the number of internal nodes of the structure. In its simplest form, this objective, which we will henceforth refer to as CECO, can then be defined as:

\begin{align}
  \mathcal{L}_\text{CECO} =  \frac{1}{2} (\mathcal{L}_{CE} + \mathcal{L}_{cont}) 
\end{align}

\section{Experiments}

\subsection{Setup}
\label{subsec: setup}
For all experiments, we utilise a pre-training set of $\approx$10 million tokens. We make this choice because Self-StrAE is intended to be data efficient, especially if it is to be useful for low resource languages where scale may not be available. For English the data was sourced from a subset of Wikipedia, while for Afrikaans and Spanish we obtained corpora from Leipzig Corpora Collection\footnote{For both Spanish and Afrikaans we selected the mixed corpus and took a uniform subsample to reduce size to the requisite scale.}. We utilise a pre-trained BPE tokenizer for each language from the BPEMB Python package \cite{bpemb}. Though the package also provides pre-trained embeddings, we solely use the tokenizer and learn embeddings from scratch. 

During the course of model development, we utilised additional evaluation sets as a further guide. For English, we used Simlex \cite{simlex} and Wordsim353 \cite{Wordsim353} as measures of how well the model captures lexical semantics, and STS-12 \cite{sts2012}, STS-16 \cite{sts2016} and STS-B \cite{STS-B}. For Afrikaans, due to lack of resources, we utilised a Dutch translation of STS-B \cite{msts} as the two languages are closely related. For Spanish, we utilised a Spanish translation of STS-B from the same source, as well as the labelled train and dev sets from SemRel 2024 \cite{semrel}. While these sets contain labels, we apply the model fully unsupervised and solely use them for zeroshot evaluation. 

We train Self-StrAE for 15 epochs using the Adam optimizer at a learning rate of 1e-3 \cite{adam}. We set the embedding dimension to 256, with a batch size of 512 and $\tau$ of 1.2. We conducted our primary experiments on English and then applied the same system design to Spanish and Afrikaans. 

\begin{figure}[t!]
  \centering
  \includegraphics[width=\columnwidth]{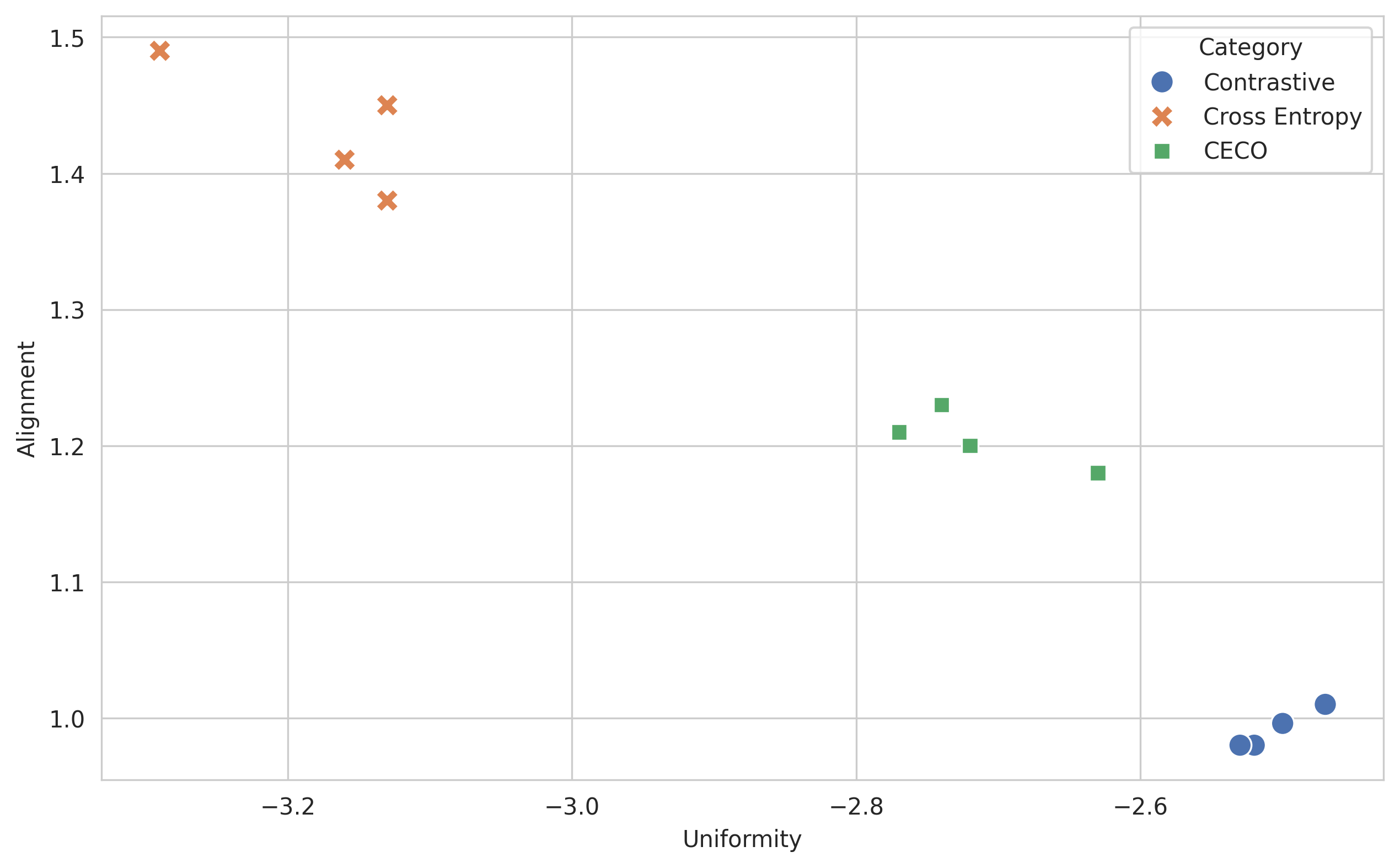}
  \caption{Uniformity and Alignment plot for \textcolor{blue}{contrastive}, \textcolor{orange}{cross entropy} and \textcolor{teal}{CECO} pre-training objectives. Results taken across four random seeds. Lower is better for both measures.}
  \label{fig:au_plots}
\end{figure}

\subsection{Which Objective is Best?}

\begin{table*}[ht!]
\centering
\resizebox{\textwidth}{!}{%
\begin{tabular}{@{}lccccccc@{}}
\toprule
Objective & Simlex & Wordsim S & Wordsim R & STS-12 & STS-16 & STS-B & SemRel (Dev) \\ \midrule
Contrastive & 13.80 $\pm$ 0.41 & 54.33 $\pm$ 0.78 & 52.40 $\pm$ 0.87 & 31.93 $\pm$ 1.03 & 52.48 $\pm$ 0.44 & 40.05 $\pm$ 2.01 & 50.13 $\pm$ 0.88 \\
CE & 13.77 $\pm$ 9.43 & 46.43 $\pm$ 24.00 & 51.23 $\pm$ 23.04 & 17.68 $\pm$ 4.88 & 25.40 $\pm$ 15.60 & 22.43 $\pm$ 15.12 & 32.95 $\pm$ 14.93 \\
CECO & \textbf{19.15 $\pm$ 2.39} & \textbf{58.33 $\pm$ 3.31} & \textbf{62.65 $\pm$ 2.76} & \textbf{41.20 $\pm$ 4.04} & \textbf{58.40 $\pm$ 1.35} & \textbf{48.35 $\pm$ 1.36} & \textbf{54.40 $\pm$ 0.81} \\ \bottomrule
\end{tabular}
}
\caption{Comparison of Objective Performance. Results are taken across four random intialisations. Models are trained on English.}
\label{tab: objectives}
\end{table*}

\begin{table*}[ht!]
\centering
\resizebox{\textwidth}{!}{%
\begin{tabular}{@{}lccccccccc@{}}
\toprule
\textit{k} & \textit{u} & Simlex & Wordsim S & Wordsim R & STS-12 & STS-16 & STS-B & SemRel (Dev) & \# Params \\ \midrule
8 & 32  & 17.50 $\pm$ 2.12 & 58.45 $\pm$ 1.04 & 62.10 $\pm$ 2.29 & 31.00 $\pm$ 2.67 & 52.53 $\pm$ 3.33 & 41.90 $\pm$ 2.09 & 49.30 $\pm$ 0.59 & 4192 \\
32 & 8  & 17.28 $\pm$ 5.94 & 44.83 $\pm$ 27.11 & 49.10 $\pm$ 25.47 & 33.28 $\pm$ 17.49 & 46.75 $\pm$ 30.85 & 41.35 $\pm$ 25.57 & 43.95 $\pm$ 30.50 & 280 \\
64 & 4  & 16.15 $\pm$ 9.82 & 48.63 $\pm$ 20.95 & 51.30 $\pm$ 23.05 & 38.88 $\pm$ 22.39 & 49.48 $\pm$ 31.05 & 43.05 $\pm$ 28.91 & 46.13 $\pm$ 30.35 & 88 \\
128 & 2  & 17.33 $\pm$ 7.12 & 52.85 $\pm$ 19.33 & 55.15 $\pm$ 19.85 & 39.63 $\pm$ 20.83 & 50.38 $\pm$ 31.92 & 46.63 $\pm$ 27.95 & 47.78 $\pm$ 30.92 & 22 \\
256 & 1  & 12.00 $\pm$ 12.84 & 42.80 $\pm$ 23.35 & 45.05 $\pm$ 24.58 & 29.18 $\pm$ 24.68 & 39.65 $\pm$ 32.22 & 37.35 $\pm$ 29.55 & 40.63 $\pm$ 29.07 & 7 \\ \midrule
8 & 32  & 19.4 & 59.4 & 64.3 & 27.6 & 56 & 44.5 & 50.1 & 4192\\
32 & 8 & 21.6 & 57.2 & 61.6 & 44.3 & 63.3 & 54.1 & 58.8 & 280 \\
64 & 4  & \textbf{21.7} & 62.8 & 66.1 & 49.9 & 65.6 & 57.4 & 61.3 & 88 \\
128 & 2  & 18.4 & \textbf{65.1} & \textbf{67.2} & 49 & \textbf{67.2} & 60.9 & 63.2 & 22 \\
256 & 1  & 20.7 & 63.2 & 66.3 & \textbf{50.1} & 66.2 & \textbf{61.6} & \textbf{63.6}& 7 \\ 
\bottomrule
\end{tabular}
}
\caption{Impact of number of independent channels on performance. Results are taken across four random initialisations. Models are trained on English. Top half of the table represents average performance, the bottom half contains the best performing initialisation. \# Params is the number of non-embedding parameters.}
\label{tab: channels}
\end{table*}

The first thing we want to establish is which objective is most suitable for training Self-StrAE, as the original version only utilises contrastive loss. For parity with the original implementation, we treat the embeddings as square matrices (i.e. \textit{k} = \textit{u}) in this experiment. 

Figure \ref{fig:au_plots} show the uniformity and alignment analysis \cite{ua} of the representations learned by the different objectives. Uniformity describes the extent to which embeddings are spread around the space, while alignment characterises how similar positive target pairs are to each other. To be successful, representations should optimise both properties. We can observe that while the cross entropy objective leads to uniformity, it is comparatively poor at optimising alignment. This essentially implies that the decoder embeddings deviate from those of the encoder. Alignment is clearly a desireable property, as the results in table \ref{tab: objectives} show. The contrastive loss leads to both better sentence level representations and to more stable performance. 

However, the best setting of all is CECO (the combination of cross entropy and contrastive). There are two factors worth considering that may explain this finding. Firstly, including reconstruction of discrete labels inherently provides additional meaningful information compared to just organising the representations alone. Secondly, at the token level the contrastive loss is most susceptible to noise (e.g. the word 'the' may occur frequently in the batch, but each repeated instance will be treated as a false negative), and under such conditions the objective has been shown to lead to feature suppression \cite{cshort}. 

\noindent\textbf{Summary:} We find that combining cross entropy and contrastive loss leads to better representations than applying each objective individually, and consequently use this approach going forward.

\subsection{How many channels?}

Each embedding in Self-StrAE is treated as consisting of \textit{k} independent channels of size \textit{u}. This is intended to allow the representations to capture different senses of meaning. However, in the original paper the number of channels is set to be the square root of \textit{u}, and not explored further. Consequently, we wanted to see what the optimal balance between the number of channels and their size was. Results are shown in \ref{tab: channels}. Surprisingly, we found that as the number of channels increased (and consequently \textit{u} decreased) performance improves quite dramatically, even to the limit of treating each value in the embedding as independent. Furthermore, because the number of non-embedding parameters (i.e. the composition and decomposition functions) is directly tied to the channel size \textit{u}, \textit{decreasing model complexity improves embedding quality}. 

However, it should be noted that this decrease in complexity comes with a tradeoff in terms of reliability. The smaller the size of the channel, the more variance we observed between random initialisations, with some initialisations failing to learn any meaningful representations whatsoever. We have found a solution that is able to maintain performance and ensure stability between seeds, but we leave discussion of this to the appendix, as we do not yet have a clear picture of what exactly is causing instability and wish to avoid speculation. We do however wish to emphasise that the problem is tractable and there is ample scope for further development, and direct the interested reader to \ref{sec:appendix} for more information. 

\noindent\textbf{Summary:} Increasing the number of channels while decreasing their size leads to significant improvements in performance, though at the cost of some instability between seeds. For our submisson to SemRel we used the setting \textit{k} = 128, \textit{u} = 2 as this allowed for an acceptable failure rate while not compromising performance (roughly 1 in 4 seeds fail). Consequently, our system utilises only 22 non-embedding parameters.

\begin{table*}[ht!]
\centering
\resizebox{\textwidth}{!}{%
\begin{tabular}{@{}lccccc@{}}
\toprule
Language & NL STS-B (Dev) & NL STS-B (Test) & Afr SemRel (Dev) & Afr SemRel (Test) & Competition Rank \\ 
Afrikaans  & 52.8 & 64.5 & 23.4 & 76.5 & 2 \\ \midrule
Language & ESP STS-B (Test) & ESP SemRel (Train) & ESP SemRel (Dev) & ESP SemRel (Test) & Competition Rank \\ 
Spanish & 61.5 & 58.5 & 68.7 & 63.5 & 6 \\ \bottomrule
\end{tabular}%
}
\caption{Self-StrAE Performance on Spanish and Afrikaans. Results correspond to those of the submitted systems, which we selected using the best run from four random initialisations.}
\label{tab: ml}
\end{table*}
\subsection{Performance Across Languages}
So far our experiments have only considered English. We now examine whether the framework is language agonistic, and pre-train Self-StrAE on both Spanish and Afrikaans. As before we pretrain on a small scale data (described in \ref{subsec: setup}).

Results are in \ref{tab: ml}. We can see that the improvements to Self-StrAE hold across different languages and are not the result of some quirk in our English pre-training set. In fact performance is either comparable or better than on English. The results on Afrikaans are particularly interesting as the model performs significantly better on this language. Whether this is due to how the test set was created or to underlying features of the language provides an interesting question for future work. Moreover, the Afrikaans model, despite never having been trained on Dutch, is able to generalise fairly well to it, shown by the results on the translated STS-B sets.  

\section{Related Work}

\noindent\textbf{Recursive Neural Networks:} Self-StrAE belongs to the class of recursive neural networks first popularised by \cite{socher-2011, socher-2013}. Recursive neural networks are extremely similar to recurrent neural networks, they differ because they process inputs hierarchically rather than sequentially (e.g. going up a parse tree).

\noindent\textbf{Learning Structure and Representations:} Recursive neural networks require structure as input. An alternative approach is to train a model that learns structure and the network at the same time. Recent unsupervised examples include \citet{diora, s-diora, r2d2}. However, these mechanisms generally use search to determine structure making them highly memory intensive. Self-StrAE differs from these as it asks the representations to define their own structure, making it much more resource efficient, though less flexible in certain aspects. 

\noindent\textbf{Contrastive Loss:} Contrastive loss is an objective which optimises the representation space directly. In broad terms this objective requires the representations of a positive pair to be as similar to each other as possible, while minimising similarity to a set of negative examples. The closest examples of this objective, for the approach employed in this paper, are \citet{simclr, jimmy, CLIP}.

\section{Conclusion}
We show that two simple changes can make Self-StrAE significantly more performant: adding a discrete reconstruction objective and increasing the number of independent channels. The latter also has the added benefit of reducing the number of parameters in the model, and surprisingly means that simpler is better. More broadly, we believe these findings demonstrate the potential of an inductive biases towards explicit structure. Self-StrAE, at present, is a very simple model. The only thing it really has going for it is the inductive bias which tasks embeddings with organising themselves hierarchically. While the gap between Self-StrAE and SoTA systems still remains, the fact that it is able to perform at all demonstrates the promise. Moreover, the fact that the two simple changes demonstrated in this paper can lead to such improvements indicates that the full potential of the inductive bias has yet to be reached, and it is likely that further refinements can lead to even more substantial benefits. Finally, because this model does not require significant scale to optimise pursuing further improvements may provide substantial benefits for low resource languages where pre-training data is scarce. 

\section{Limitations}
The results in this paper represent steps towards an improved model rather than a complete picture. We still do not fully understand what causes the instability in training when the number of channels increased, and though we can provide a solution (see \ref{sec:appendix}), further analysis is needed. The performance of contrastive loss can depend quite heavily on how positive and negative examples are defined and it is likely that the explanation rests there. Secondly, while we have shown that Self-StrAE can be applied to languages other than English the results are limited to Indo-European languages. An interesting avenue for future work would be investigating a broader spectrum of languages, and whether specific characteristics can be identified which influence how well the model performs. 

\clearpage 

\section{Acknowledgements}
MO was funded by a PhD studentship through Huawei-Edinburgh Research Lab Project 10410153.
We thank Victor Prokhorov, Ivan Vegner and Vivek Iyer for their valuable comments and helpful suggestions during the creation of this work.



\bibliography{custom}

\appendix

\section{Stabilising High Channel Self-StrAE}
\label{sec:appendix}
One solution we have found to the instability issue is modifying the objective. This formulation, loosely inspired by SimCSE \cite{simcse}, runs the same input through the model twice, with different dropout masks applied each time. The objective is cross entropy reconstruction for the leaves, and contrastive loss between the two different sets of decoder embeddings for the non-terminals. Currently we have two theories as to why this might work:

\begin{itemize}
    \item Better negatives: because the decoder embeddings represent the contextualised meaning of node rather than it's local one, the issue of false negatives is somewhat mitigated. 
    \item Encoder consistency: because we ask the two sets of decoder embeddings to be similar to each other the encoder is encouraged to produce the same structure regardless of dropout mask. It may be that this pressure towards regularity leads to the improved consistency. 
\end{itemize}

Results are shown in \ref{tab: alt}. For lack of a better term we refer to this alternative objective as StrCSE. In its current form we do not consider this objective to be well formed, and solely provide it here as a possible starting point for further research. 

\begin{table*}[h!]
\centering
\resizebox{\textwidth}{!}{%
\begin{tabular}{@{}lccccccc@{}}
\toprule
Objective & Simlex & Wordsim S & Wordsim R & STS-12 & STS-16 & STS-B & SemRel (Dev) \\ \midrule
Contrastive & 13.80 $\pm$ 0.41 & 54.33 $\pm$ 0.78 & 52.40 $\pm$ 0.87 & 31.93 $\pm$ 1.03 & 52.48 $\pm$ 0.44 & 40.05 $\pm$ 2.01 & 50.13 $\pm$ 0.88 \\
CE & 13.77 $\pm$ 9.43 & 46.43 $\pm$ 24.00 & 51.23 $\pm$ 23.04 & 17.68 $\pm$ 4.88 & 25.40 $\pm$ 15.60 & 22.43 $\pm$ 15.12 & 32.95 $\pm$ 14.93 \\
CECO & \textbf{19.15 $\pm$ 2.39} & \textbf{58.33 $\pm$ 3.31} & \textbf{62.65 $\pm$ 2.76} & 41.20 $\pm$ 4.04 & 58.40 $\pm$ 1.35 & 48.35 $\pm$ 1.36 & 54.40 $\pm$ 0.81 \\ 
CECO k=128 u=2 & 17.33 $\pm$ 7.12 & 52.85 $\pm$ 19.33 & 55.15 $\pm$ 19.85 & 39.63 $\pm$ 20.83 & 50.38 $\pm$ 31.92 & 46.63 $\pm$ 27.95 & 47.78 $\pm$ 30.92 \\
StrCSE k=128 u=2 & \textbf{21.68 $\pm$ 1.88} & \textbf{59.06 $\pm$ 2.38} & \textbf{64.08 $\pm$ 0.91} & \textbf{49.46 $\pm$ 0.59} & \textbf{66.18 $\pm$ 0.24} & \textbf{61.30 $\pm$ 0.76} & \textbf{62.88 $\pm$ 0.42} \\
\bottomrule
\end{tabular}
}
\caption{StrCSE compared with other objectives. Results are taken over four random intialisations. Training data is English.}
\label{tab: alt}
\end{table*}

\end{document}